\documentclass[journal,twoside,web]{ieeecolor}

\usepackage{chapterbib}
\usepackage{graphicx}%
\usepackage{multirow, multicol}%
\usepackage{amsmath,amssymb,amsfonts}%
\usepackage{mathrsfs}%
\usepackage{makecell}
\usepackage{xcolor}%
\usepackage{textcomp}%
\usepackage{manyfoot}%
\usepackage{booktabs}%
\usepackage{algorithm}%
\usepackage{algorithmicx}%
\usepackage{algpseudocode}%
\usepackage{longtable}
\usepackage{lscape}
\usepackage{listings}%
\usepackage{epsfig} 
\DeclareMathOperator{\Tr}{Tr}
\usepackage{breqn}
\usepackage{bm}
\usepackage{float}
\usepackage{siunitx}
\usepackage{soul}
\usepackage{hyperref}
\usepackage{pdfpages}
\usepackage{subfiles}
\usepackage{generic}
\usepackage{cite}
\usepackage{graphicx}
\usepackage{textcomp}

\IEEEaftertitletext{\vspace{-30pt}}

\DeclareGraphicsExtensions{.png,.pdf}
\makeatletter
\newcommand*{\rom}[1]{\expandafter\@slowromancap\romannumeral #1@}

\usepackage{textcomp}
\def\BibTeX{{\rm B\kern-.05em{\sc i\kern-.025em b}\kern-.08em
    T\kern-.1667em\lower.7ex\hbox{E}\kern-.125emX}}
\begin{document}

\title{Beyond Humanoid Prosthetic Hands:\\Modular Terminal Devices That Improve\\User Performance}
\author{
Digby Chappell,
Barry Mulvey,
Shehara Perera,
Fernando Bello,
Petar Kormushev,
Nicolas Rojas
\thanks{This work was supported in part by the UKRI CDT in AI for Healthcare under Grant No. EP/S023283/1.
(\textit{Corresponding author: Digby Chappell, email:
\href{mailto:dchappell@seas.harvard.edu}{dchappell@seas.harvard.edu}})
}
\thanks{All authors are with are with Imperial College London, London SW7 2BX, UK.}
\thanks{Digby Chappell is also with Harvard University, Cambridge, MA 02134, USA.}
\thanks{Shehara Perera is also with the University of Cambridge, Cambridge CB3 0FA, UK.}
\thanks{Nicolas Rojas is also with The AI Institute, Cambridge, MA 02142, US.}
}

\maketitle

\begin{abstract}
Despite decades of research and development, myoelectric prosthetic hands lack functionality and are often rejected by users. This lack in functionality can be partially attributed to the widely accepted anthropomorphic design ideology in the field; attempting to replicate human hand form and function despite severe limitations in control and sensing technology. Instead, prosthetic hands can be tailored to perform specific tasks without increasing complexity by shedding the constraints of anthropomorphism. In this paper, we develop and evaluate four open-source modular non-humanoid devices to perform the motion required to replicate human flicking motion and to twist a screwdriver, and the functionality required to pick and place flat objects and to cut paper. Experimental results from these devices demonstrate that, versus a humanoid prosthesis, non-humanoid prosthesis design dramatically improves task performance, reduces user compensatory movement, and reduces task load. Case studies with two end users demonstrate the translational benefits of this research. We found that special attention should be paid to monitoring end-user task load to ensure positive rehabilitation outcomes.
\end{abstract}


\section{Introduction}\label{sec:intro}

\begin{figure}[!t]
    \centering
    \includegraphics[width=0.95\columnwidth]{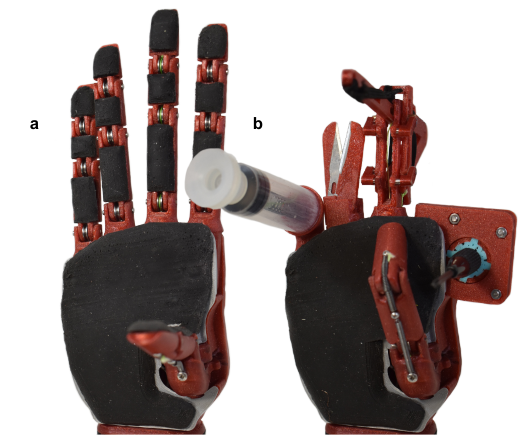}
    \vspace{-6pt}
    \caption{\textbf{a} Conventional humanoid myoelectric prosthesis. \textbf{b} Modular, non-humanoid terminal devices mounted on the myoelectric prosthesis base platform.
    }
    \label{fig:main}
\end{figure}

\IEEEPARstart{M}{illions}
of adults worldwide are affected by upper limb difference (ULD)~\cite{McDonald2021GlobalAmputation}---acquired either congenitally (from birth) or later in life through amputation. For many, myoelectric prosthetic hands using non-invasive electromyography for control offer the possibility of restoring some level of capability. Myoelectric prostheses utilise the small electrical signals produced in a user's residual limb during muscle contraction to enable volitional control of an artificial hand, such as the prosthetic hand shown in Fig.~\ref{fig:main}~\textbf{a}, or other device---promising a more ergonomic and versatile solution than body-powered prostheses, and in the case of non-invasive electromyography, a less invasive and costly solution than myoelectric prostheses relying on implanted electrodes.

Myoelectric prosthetic hand design has been the subject of decades of research, beginning in the 1950s with simple, single degree of freedom (DOF) devices~\cite{Battye1955TheProstheses.}.
In the years since, advances in hardware have allowed multiple degrees of freedom (DOFs) to be actuated independently~\cite{Kyberd1994TheProsthesis., Weiner2018TheControl}, promising to dramatically improve the capabilities of such prostheses.
However, this has not been the case: existing multi-articulating myoelectric hands demonstrate little-to-no physical benefit over simpler, single action prostheses~\cite{Kerver2023ThePromise}.
Despite the introduction of multi-articulating prosthetic hands, user dissatisfaction and rejection rates have remained high~\cite{Kyberd2007SurveyKingdom, Pylatiuk2007ResultsUsers, stlie2012ProsthesisSurvey, Salminger2020CurrentAcceptance}, with a lack of functionality and dexterity often cited as key reasons for this~\cite{Biddiss2009ConsumerProsthetics, Espinosa2019UnderstandingAbandonment, Stephens-Fripp2019AHands, Yamamoto2019Cross-sectionalProstheses}.
This poor functionality means that adults with ULD are often forced to either change profession or adapt their duties in order to continue working~\cite{Datta2004FunctionalCongenital, vanderSluis2009JobAmputees, Postema2016UpperProductivity}.
It is clear that there is a shortfall between current developments and users' needs.

The majority of myoelectric prosthetic hands are anthropomorphic (humanoid), in that they seek to imitate the form and function of human hands.
However, they fail to meet this goal due to two reasons: mechanical limitations and control limitations.
Mechanically, there is a limit on the number of actuators that can be housed within the form factor of a human hand. This means that the majority of anthropmorphic prostheses have between $5$ and $7$ DoFs that can be independently actuated~\cite{Kyberd1994TheProsthesis., Liow2020OLYMPIC:Mechanisms, Johannes2020TheLimb}.
From a control perspective, myoelectric controllers for anthropomorphic prostheses are currently limited to controlling between $1$ and $3$ independent DoFs simultaneously~\cite{Hahne2018_LinearRegression2DOF, Dyson2018MyoelectricDecoders, Dyson2020LearningControl}.
Given these limitations, it may be the case that non-humanoid, functional designs of upper limb prostheses could be favourable alternatives to existing humanoid prostheses.

Functional, non-humanoid upper limb myoelectric prostheses are rare.
Existing examples of non-humanoid myoelectric prostheses are grippers~\cite{Dollar2007TheStudy, Cheng2016ProstheticRobotics, Yoshikawa2023Finch:Bulge, Hong2023Angle-programmedUltraprecision}, seeking to improve on the existing grasping functionality of prosthetic hands.
Looking beyond myoelectric prostheses, functional designs are {more often seen} in body-powered and static prostheses, where even greater limitations on mechanical and control complexity exist.
{Non-humanoid body-powered and static prostheses have been in use for over a century to achieve tasks that are impossible with conventional humanoid designs~\cite{Kyberd2021MakingArms}.
In modern times, a variety of non-humanoid static devices have been developed~\cite{Highsmith2007KinematicAmputation, Hammond2012TowardsHands, Maat2018PassiveReview}, with commercial options such as the Nicole ALX (Koalaa Ltd; UK) and the TRS Helix (Therapeutic Recreation Systems Inc, Fillauer; USA) widely available.
Modern body-powered prostheses, however, are less varied; the split-hook design is extremely popular for general use as a gripper~\cite{Smit2010EfficiencyProstheses, Smit2012EfficiencyDevelopment}, but few alternative designs for task-specific body-powered prostheses exist.}
One notable commercial example is the Ski-2 Terminal Device (Therapeutic Recreation Systems Inc, Fillauer; USA), which allows the user to adjust the angle of a ski pole using a body-powered cable.
There is a scarcity of research evaluating the performance of functional prostheses~\cite{Bragaru2012SportLiterature}, but the commercial popularity of task-specific passive and body-powered variants provides motivation for the wider study of task-specific designs of upper limb prosthetic devices.
Myoelectric prostheses specifically, while still limited, can leverage more complex mechanisms and controllers than body-powered or passive prostheses.
Functional, non-humanoid designs of myoelectric prosthetic hands may therefore hold promise as an area of research.

Task-specific upper limb prostheses bring forth unique challenges.
While functionally beneficial for individual tasks, the specificity of task-specific devices limit their utility in daily life.
The ability to change end effector may be needed in order for task-specific prostheses to benefit users functionality day to day.
Furthermore, an anthropomorphic appearance is often desired~\cite{Walker2020TowardsDifference}, and contributes positively to social acceptance~\cite{Arabian2016GlobalDevices, Biddiss2009ConsumerProsthetics}.
It therefore follows that users should be able to obtain the functional benefits of non-humanoid terminal devices while retaining the option of an humanoid prosthesis when desired --- something that can be achieved through designed modularity.
Several examples of modular myoelectric hands exist in literature~\cite{Jiang2014AHand, Johannes2020TheLimb}, including the OLYMPIC hand used in this study~\cite{Liow2020OLYMPIC:Mechanisms}.
However, modularity in existing hands has so far been leveraged for adapting to varying levels of ULD~\cite{Johannes2020TheLimb} and for component replacement and repair~\cite{Jiang2014AHand, Liow2020OLYMPIC:Mechanisms}.
Modularity in combination with task-specific end effectors for myoelectric prostheses remains unexplored.

In this work, we present and evaluate a series of four non-humanoid modular terminal devices that each replace one of the fingers of the OLYMPIC hand~\cite{Liow2020OLYMPIC:Mechanisms}.
The four devices, shown mounted on the base prosthesis in Fig.~\ref{fig:main}, are designed to perform actions that have meaningful social and functional uses in many areas of daily life, but are difficult-to-impossible with conventional prostheses{: flicking, screwdriving, picking and placing small flat objects, and using scissors.
With a} {humanoid manipulator, each of these tasks requires the coordination of multiple joint-space DoFs to achieve a single task-space DoF output.}
Through a control group study with participants without ULD, the four non-humanoid terminal devices were evaluated against a humanoid prosthesis on their specific tasks in terms of task performance, compensatory motion, and perceived task load.
Finally, validation case studies with two participants with ULD were performed to confirm that benefits successfully translate to real end users.

\section{Non-Humanoid Terminal Devices}
\begin{figure}[!t]
    \centering
    \includegraphics[width=0.7\columnwidth]{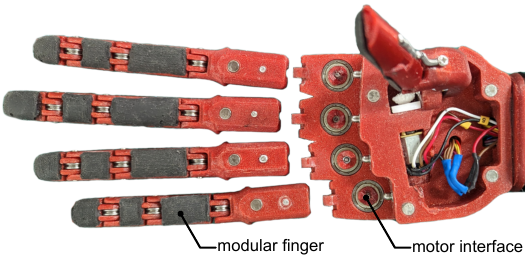}
    \vspace{-6pt}
    \caption{The OLYMPIC hand base platform, showing motor interfaces and modular fingers. Each terminal device replaces one finger.
    }
    \label{fig:olympic}
\end{figure}

We present four terminal devices that serve as illustrative, first-of-their-kind examples of task-specific myoelectric prosthetic end effectors. Each device is modular, replacing one finger of the OLYMPIC hand (see Fig. \ref{fig:olympic}), making use of tool-less finger modularity for easy replacement~\cite{Liow2020OLYMPIC:Mechanisms}. The devices are designed to perform tasks that are difficult for conventional humanoid prosthetic hands: flicking, twisting, picking and placing flat objects, and cutting. Each of these tasks require complex coordination at the joint level of a humanoid hand in order to produce a relatively simple task-space output.
All devices were manufactured using 3D printing and readily available, low-cost components. The CAD files of the OLYMPIC hand and all terminal devices are open-source and available at: \url{https://sites.google.com/view/non-human-hands}.
The surface EMG sensing Myo Armband (Thalmic Labs; USA) was used for device control. Only two electrodes were used, located on the anterior and posterior forearm, corresponding to the bulk muscles that control wrist flexion and extension, respectively. For participant $2$ with ULD, the partially reinnervated muscles of the upper arm were used (see Table \ref{tab:sup_uld_participants}). The Myo Armband samples EMG signals at $200$~Hz, and the mean absolute value $s(k)$ of the raw signal $e(k)$ from each electrode at sample $k$ was computed with a rolling window of $W=20$ samples:
\begin{equation}
    s(k) = \frac{1}{W}\sum_{w=1}^{W}e(k+w-W)
\end{equation}
The remainder of this section details the design and control of each terminal device, as well as the modular base platform on which each device is mounted.

\subsection{The OLYMPIC Hand as a Test Platform}\label{subsec:olympic}
The OLYMPIC hand is a modular prosthesis, where each finger of the hand can be removed without the need for external tools~\cite{Liow2020OLYMPIC:Mechanisms}. When a finger is removed, a corresponding motor interface is exposed (see Fig.~\ref{fig:olympic}). The terminal devices presented in this paper retarget one of these motor DOFs to perform specific tasks. The motor interfaces for the fingers are all in-plane, simplifying terminal device interface and transmission design considerably. During evaluation of each non-humanoid terminal device, the index finger of the OLYMPIC hand was replaced with the device, and the remaining fingers driven to their fully closed positions.

\begin{figure}[!t]
    \centering
    \includegraphics[width=0.8\columnwidth]{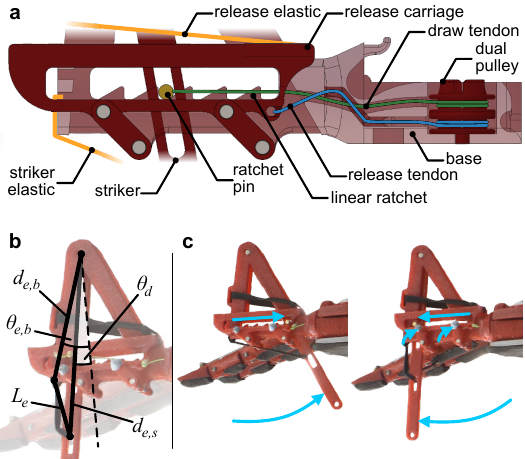}
    \vspace{-6pt}
    \caption{The flicking device. \textbf{a} Labelled section view showing tendon routing. \textbf{b} Schematic diagram showing variables used in striker elastic length calculation. \textbf{c} The flicking device in operation, from left to right: rest state, drawn state, release state.
    }
    \label{fig:flicking}
\end{figure}

\subsection{Terminal Device 1: Flicking Motion}\label{subsec:flicking}
The ability to produce a flicking motion is one that is used in social settings such as flipping a coin or taking part in traditional games such as carrom, shove ha'penny, or paper football. Flicking forms a part of every day life and has been overlooked by prosthesis designers to date. Flicking is the sudden release of stored mechanical energy; although many prostheses do use elastic elements which store energy, the execution of a flicking motion is difficult as it requires the coordination to block and release the extension of one finger. The flicking device presented in this work is able to produce a flicking motion using a single motor input.

The flicking device, shown in Fig.~\ref{fig:flicking}, uses two tendons. The first is used to load the `striker' by drawing a pin along a linear ratchet and storing elastic energy. The second tendon releases the striker by raising the pin above the ratchet, where it can move freely. Both tendons are wound around the same `dual' pulley in opposite directions; when the draw tendon is in tension, the release tendon is slack, and vice versa. As the release tendon winds to remove slack then actuate the release carriage, the draw tendon accumulates excess slack tendon length such that, when the pin is released, the striker can move along its full trajectory unhindered. The ratchet enables the striker to be loaded to $5$ levels of energy storage, enabling modulation of the impulse transferred to the flicked object.

The striker arm has a length of $L_s=125.0$~mm, and the pin moves along a line that is $d_h=53.8$~mm vertically below the axis of rotation of the striker, bringing the striker arm from a rest angle of $\theta_{\text{min}}=-5.0^\circ$ to a maximum draw of $\theta_{\text{max}}=30.0^\circ$. Given a pin position of $x$ mm from rest, the draw angle can be calculated:
\begin{equation}
    \theta_d = \text{atan2}(x+x_0,\, d_h),
\end{equation}
where $x_0$ is equal to $d_h\tan(\theta_{\text{min}})$. The elastic cord that stores energy is attached at a distance from the striker base of $d_{e,b}=62.4$~mm and angle $\theta_{e,b}=-11.7^\circ$ from vertical, and at the other end to the striker at a distance of $d_{e,s}=85.0$~mm from the striker pin. The elastic cord length $L_e$ is therefore:
\begin{equation}
    L_e = \sqrt{d_{e,s}^2 + d_{e,b}^2 - 2d_{e,s} d_{e,b}\cos(\theta_d - \theta_{e,b})},
\end{equation}
giving a tendon force of $F_e=K(L_e-L_0)$, where $K$ and $L_0$ are the spring constant and relaxed length of the elastic cord, respectively. Finally, the striker torque is equal to:
\begin{equation}
    \tau_s = F_e d_{e,s} \sin(\theta_{e,s}),
\end{equation}
where $\theta_{e,s}$ is the angle that the cord makes with the striker, calculated using the sine rule: $\sin(\theta_{e,s})=\frac{d_{e,b}}{L_e}\sin(\theta_d - \theta_{e,b})$.

The commanded position of the motor that draws and releases the striker, $q_{\text{flick}}$, was controlled by the difference between the mean absolute value signals from the two electrodes, $s_1(t)$ and $s_2(t)$. During drawing, the motor position was incremented according to a constant draw rate of $\bar{v}$. To release, the motor was driven directly to its release position $q_{\text{release}}$, which raises the release carriage:
\begin{equation}
    q_{\text{flick}}(t+\Delta t) = \begin{cases}
        q_{\text{flick}}(t) + \bar{v} \Delta t & s_1(t) - s_2(t) > \bar{s}_{\text{draw}} \\
        q_{\text{release}} & s_2(t) - s_1(t) > \bar{s}_{\text{release}} \\
        q_{\text{flick}}(t) & \text{otherwise}
    \end{cases},
\end{equation}
where $\bar{s}_{\text{draw}}$ and $\bar{s}_{\text{release}}$ are activation thresholds, and $\Delta t$ is the control rate of the prosthesis.

\begin{figure}[!t]
    \centering
    \includegraphics[width=0.8\columnwidth]{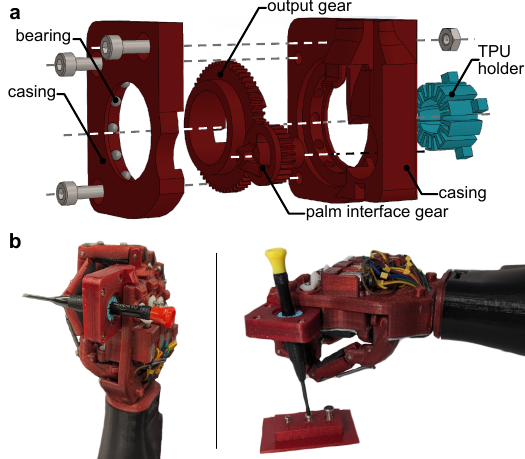}
    \vspace{-6pt}
    \caption{The twisting device. \textbf{a} Labelled exploded view. \textbf{b} The twisting device in operation, left: a screwdriver inserted into the device, right: the device being used to produce the twisting motion with a screwdriver.
    }
    \label{fig:twisting}
\end{figure}

\subsection{Terminal Device 2: Twisting Screwdriver Motion}\label{subsec:screwdriver}
The second terminal device produces the twisting motion required to use a screwdriver. Coordinating the fingers of an humanoid prosthesis to rotate the fingertips around a remote centre of motion is extremely difficult. For a non-humanoid robot, however, this can be achieved in a number of simple ways, such as with a gear, belt, or screw drive. Shown in Fig.~\ref{fig:twisting}, the twisting device uses a gear drive with gear ratio $1:2.4$ ($12.5$~mm $:$ $30$~mm), chosen to take advantage of the rest `face-down' orientation of the palm of the prosthesis, meaning the twisting axis is coplanar to the motor interfaces on the hand. An interchangeable TPU holder was included in the device in order to grip and conform to the shape of the held screwdriver. The maximum screwdriver diameter that can be accommodated by the output gear is $17.5$~mm.

In order to control this terminal device, a simple EMG controller based on thresholding was implemented. The commanded torque of the corresponding motor is controlled by the difference between the smoothed signals from the two electrodes, $s_1(t)$ and $s_2(t)$. When the differential signal $s_1(t)-s_2(t)$ rises above a threshold $\bar{s}$, the terminal device motor is driven at maximum torque $\tau_{\text{max}}$. Similarly, when the negative differential signal $s_2(t)-s_1(t)$ rises above the same threshold, the motor is driven in reverse:
\begin{equation}
    \tau_{\text{screwdriver}} =
    \begin{cases}
        \tau_{\text{max}} & s_1(t) - s_2(t) > \bar{s}\\
        -\tau_{\text{max}} & s_2(t) - s_1(t) > \bar{s}\\
        0 & \text{otherwise}.
    \end{cases}
\end{equation}

\subsection{Terminal Device 3: Grasping Small, Flat Objects}\label{subsec:small_flat}
The third terminal device is designed to grasp small, flat objects. The human hand is adept at performing this function, thanks to a complex array of mechanoreceptors that elicit rapid local motor responses during precision manipulation to ensure grasped objects remain held~\cite{Johansson1987SignalsGrip}. Replicating this in an anthropomorphic form is extremely challenging given sensing, actuation, and control limitations, and existing prostheses struggle considerably with precision tasks~\cite{Chappell2022TowardsDistance, Resnik2018HowProstheses, Mohammadi2022PreliminaryProsthesis}.

\begin{figure}[!t]
    \centering
    \includegraphics[width=0.8\columnwidth]{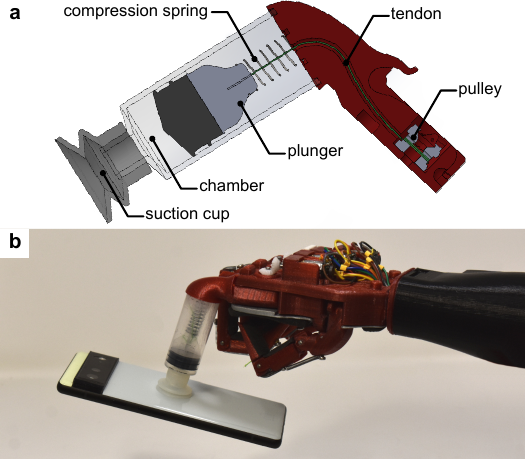}
    \vspace{-6pt}
    \caption{The suction device. \textbf{a} Labelled section view showing tendon routing. \textbf{b} The suction device in operation being used to pick up a phone.
    }
    \label{fig:suction}
\end{figure}

This terminal device utilises a suction cup controlled by a tendon-spring actuated plunger. Shown in Fig. \ref{fig:suction}, a tendon routed from the palm interface pulley to the plunger is drawn to generate a vacuum inside the chamber. As the plunger retracts, assuming that air behaves as an ideal gas, a pressure differential $\Delta P$ is created that is inversely proportional to the expanded volume of the chamber $V_1$:
\begin{equation}
    \Delta P = P_0\Big(\frac{V_0}{V_1} - 1\Big),
\end{equation}
where $P_0$ is atmospheric pressure and $V_0$ is the initial volume.

The vacuum chamber has an internal diameter of $D_v=21$~mm, and a plunger stroke of $q_{\text{max}}=25~$mm. During use, the plunger was controlled to nominally be at a rest position of $q_{\text{rest}}=5$~mm. This was because the suction cup is able to passively produce a vacuum when pressed onto the surface of an object, so the plunger must be able to produce a small positive pressure to counteract the passive vacuum and release the object. With a plunger area of $A_p=\pi D_p^2 / 4$, plunger diameter of $D_p=27$~mm, standard conditions of $P_0=1.01325$~Pa, and neglecting the plunger volume, the maximum producible suction force $F_v$ is:
\begin{equation}
    F_v = -A_pP_0\Big(\frac{q_{\text{rest}}}{q_{\text{max}} - q_{\text{rest}}} - 1 \Big),
\end{equation}
giving a maximum suction force of $43.5$~N.

Two surface EMG electrodes were used for control: the difference between the smoothed electrode signals, $s_1(t)$ and $s_2(t)$, determined the target position of the plunger:
\begin{equation}
    q_{\text{plunger}} = (0.5q_{\text{max}} - q_{\text{min}})(s_1(t) - s_2(t)) + q_{\text{rest}}.
\end{equation}

\subsection{Terminal Device 4: Cutting Paper}\label{subsec:cutting}
Cutting paper is a task typically performed with scissors, however, producing the motion required to use scissors requires each handle to be moved around a remote centre of motion, which is very difficult with existing prostheses. The final terminal device presented here achieves this same cutting functionality, but the prosthesis itself is the tool.

\begin{figure}[!t]
    \centering
    \includegraphics[width=0.8\columnwidth]{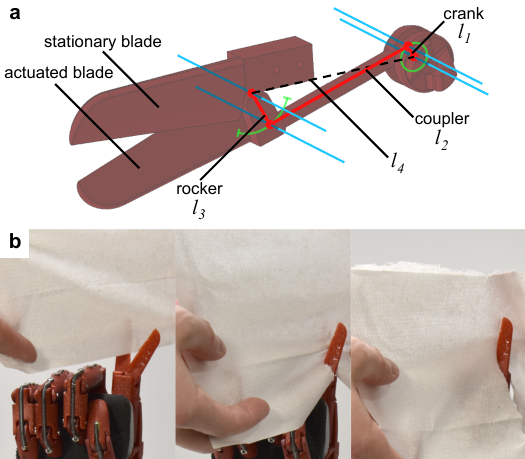}
    \vspace{-6pt}
    \caption{The cutting device. \textbf{a} Labelled internal view of the crank and rocker mechanism. \textbf{b} The cutting device in operation being used to cut a piece of tissue paper.
    }
    \label{fig:cutting}
\end{figure}

Shown in Fig. \ref{fig:cutting}, a crank and rocker mechanism is used to convert continuous rotation of the finger interface gear to oscillatory motion of one blade of the scissors, while the other blade remains stationary. With crank, coupler, and rocker link lengths of $l_1=3$~mm, $l_2=32$~mm, and $l_3=7.5$~mm, respectively, and a distance between the crank and rocker axes of $l_4=35.5$~mm, the maximum rocker angle relative to horizontal can be computed via the cosine rule:
\begin{equation}
    \theta_{\text{max}} = \arccos{\Big(\frac{l_3^2 + l_4^2 - (l_1 + l_2)^2}{2l_3l_4}\Big)},
\end{equation}
and the minimum angle:
\begin{equation}
    \theta_{\text{min}} = \arccos{\Big(\frac{l_3^2 + l_4^2 - (l_1 - l_2)^2}{2l_3l_4}\Big)},
\end{equation}
resulting in a range of motion of approximately $53.3~^\circ$.

Because the device produces continuous cutting motion, control only requires a single surface EMG electrode. When the smoothed electrode signal, $s_1(t)$, rises above a threshold $\bar{s}$, the terminal device motor is driven at maximum torque:
\begin{equation}
    \tau_{\text{scissors}} =
    \begin{cases}
        \tau_{\text{max}} & s_1(t) > \bar{s}\\
        0 & \text{otherwise}.
    \end{cases}
\end{equation}

\section{Evaluation}\label{sec:eval}

Evaluating terminal devices that do not conform to traditional prosthesis design philosophies is challenging. Existing dexterity assessments often taking the form of pick and place tasks~\cite{Jebsen1969AnFunction, Trombly1983OccupationalDysfunction, Light2002SHAP} or other simple manipulation tasks such as peg insertion~\cite{Tiffin1948TheValidity.}. For the task-specific terminal devices presented here, alternative assessments were required to extract quantifiable metrics related to task performance. However, prosthesis use goes beyond task performance, and longer-term comfort and user perception are extremely important, and therefore further metrics have been considered. In this work, as well as task performance, user perceived task load~\cite{Chappell2022VirtualInteraction, Parr2023APROS-TLX} and the compensatory motion that the user must perform with their body in order to perform a given motion~\cite{Hunt2017Pham:Measure, Vujaklija2023BiomechanicalTests} were recorded. Given that the most common reasons for prosthetic hand rejection are discomfort and weight~\cite{Salminger2020CurrentAcceptance}, monitoring compensatory motion (which can lead to longer-term discomfort) and task load---including perceived physical demand---gives us important insights into how non-humanoid terminal devices could fare during regular at-home use.

A control group study was performed to evaluate each terminal device with two groups of participants without ULD, with minimal prior experience with myoelectric prostheses; a humanoid group who used the OLYMPIC hand in its humanoid configuration, and a non-humanoid group who used the OLYMPIC hand with the non-humanoid terminal device. The characteristics of each participant group are shown in Table \ref{tab:sup_participant_groups}. A total of $64$ participants without ULD were recruited ($32$ to the non-humanoid group, $32$ to the humanoid group; $8$ participants completing an individual task per group), and each task was completed in a single session lasting approximately $1$ hour. Participants without ULD were not compensated. Following this, each device was tested with two participants with ULD as case studies, in order to verify that results observed in the control group study translated to end users (characteristics of participants with ULD are shown in Table \ref{tab:sup_uld_participants}). Both participants with ULD were not myoelectric prosthetic hand users in their daily life, but had both used myoelectric prosthetic hands in research settings before. Participants with ULD completed all tasks in a single session lasting approximately $2.5$ hours, and were not compensated aside from travel reimbursement. This study was given a favourable opinion by the Imperial College Research Governance and Integrity Team (RGIT), number 6571021, which was in accordance with the Declaration of Helsinki. Written consent was obtained from all participants prior to taking part in this study. Participants depicted gave informed consent for images to be published.

\renewcommand{\arraystretch}{1.0}
\begin{table}[t]
    \centering
    \caption{
    Characteristics of groups of participants without upper limb difference for each task.
    Age, gender, and handedness of groups of participants without upper limb difference. Standard deviation is shown in brackets.\\Groups: NH = Non-Humanoid, H = Humanoid.
    }
    \begin{tabular}{c c c c c c c}
        \hline
         &  &  & \multicolumn{2}{c}{Gender} & \multicolumn{2}{c}{Handedness} \\
        Task & Group & Age/years & Male & Female & Right & Left \\
                \hline
        \multirow{3}{*}{\makecell{Flicking\\Task}} 
        & NH  & 25.8 (4.7) & 5 & 3    & 8 & 0 \\
        & H      & 26.6 (3.2) & 5 & 3    & 7 & 1 \\
        & All           & 26.2 (3.9) & 10 & 6   & 15 & 1 \\
                \hline
        \multirow{3}{*}{\makecell{Screwing\\Task}} 
        & NH  & 23.9 (2.2) & 5 & 3    & 6 & 2 \\
        & H      & 25.1 (3.3) & 5 & 3    & 7 & 1 \\
        & All           & 24.5 (2.8) & 10 & 6   & 13 & 3 \\
                \hline
        \multirow{3}{*}{\makecell{Stacking\\Task}} 
        & NH  & 26.1 (2.7) & 5 & 3    & 6 & 2 \\
        & H      & 26.4 (2.5) & 5 & 3    & 8 & 0 \\
        & All           & 26.3 (2.5) & 10 & 6   & 14 & 2 \\
                \hline
        \multirow{3}{*}{\makecell{Cutting\\Task}} 
        & NH  & 26.1 (2.4) & 5 & 3    & 6 & 2 \\
        & H      & 26.0 (2.5) & 5 & 3    & 7 & 1 \\
        & All           & 26.1 (2.4) & 10 & 6   & 13 & 3 \\
                \hline
    \end{tabular}
    \label{tab:sup_participant_groups}
\end{table}

\begin{table*}[t]
    \centering
    \caption{
    Table of participant characteristics of individual participants with limb difference.
    For bilateral ULDs the side tested is shown in brackets, and details for the tested side are shown. Causes listed: traumatic injury (TI), planned surgery (PS). $^a$Participant $2$ acquired their ULD via planned surgery after a prior traumatic injury, and received targeted muscle reinnervation; the muscles of the upper arm were used.
    }
    \begin{tabular}{ c c c c c c c c c c c}
        \hline
        ID & Age & Gender & Affected Side & ULD Level & Cause & Years Since & Dominant Hand & Own Prosthesis & Frequency of Use \\
        \hline
        1 & 43 & Male & Both (Right) & Below Elbow & TI & 10 & Left & Body-Powered & Daily \\
        2 & 46 & Female & Right & Below Elbow$^a$ & TI (PS) & 24 (6) & Left & Static & Infrequent \\
        \hline
    \end{tabular}
    \label{tab:sup_uld_participants}
\end{table*}

\begin{figure*}[!t]
    \centering
    \includegraphics[width=0.9\textwidth]{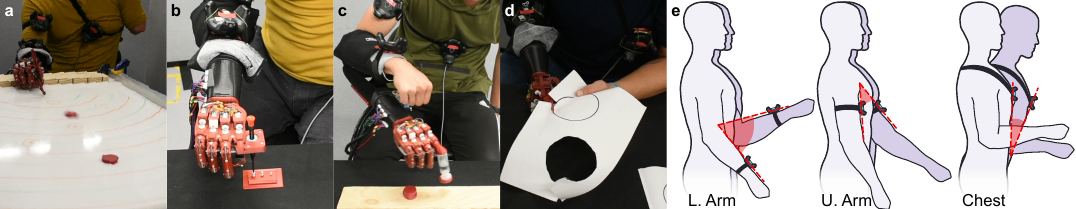}
    \vspace{-6pt}
    \caption{Evaluation techniques used in this study. \textbf{a-d} Flicking, screwing, stacking, and cutting tasks used to evaluate the performance of the flicking, twisting, suction, and cutting devices, respectively. \textbf{e} Locations of body-mounted trackers on the lower arm, upper arm, and chest used to measure mean angular deviation (shown in red).
    }
    \vspace{-10pt}
    \label{fig:evaluation}
\end{figure*}

\subsection{Task Performance}

Participants completed four tasks (shown in Fig.~\ref{fig:evaluation} a-d), each comparing a corresponding non-humanoid device with the control humanoid prosthesis. For all tasks, a single practice attempt was allowed before testing began.

\begin{figure*}[!t]
    \centering
    \includegraphics[width=180mm]{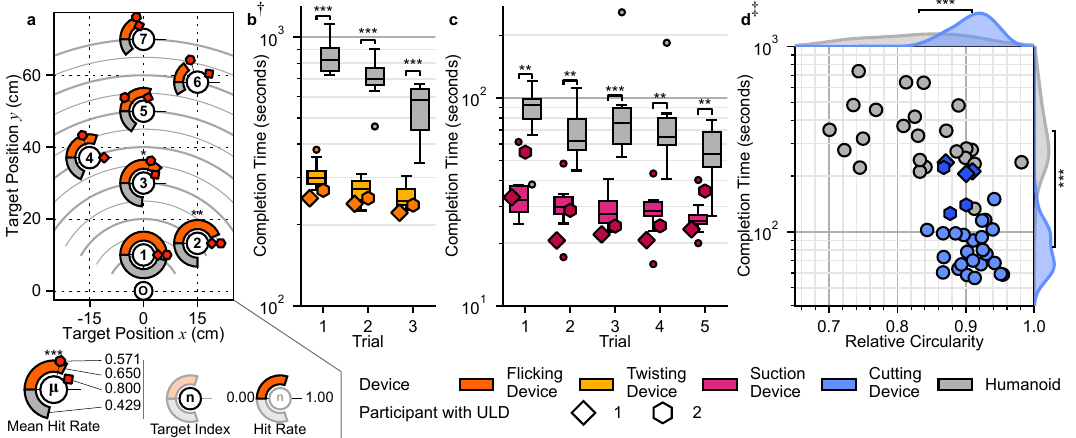}
    \vspace{-6pt}
    \caption{Task performance of each group of participants without ULD on the flicking task (\textbf{a}), screwing task (\textbf{b}), stacking task (\textbf{c}), and cutting task (\textbf{d}), with task performance of participants with ULD shown as individual points.
    \textbf{a} Scale diagram of flicked disc location ($\bigcirc$), target disc locations ($1-7$), with corresponding hit rates shown.
    \textbf{d} Scatter plot of relative circularity of the cut circle ($1.00$ indicates a perfect circle) against task completion time, with kernel density estimate plots of relative circularity and completion time shown adjacent to each axis.
    Results of (\textbf{a}) paired $z$-tests, (\textbf{c-d}) Mann-Whitney U tests: $^{*}p<0.05$, $^{**}p<0.01$, and $^{***}p<0.001$, after Bonferroni correction for $7$, $3$, $5$, and $2$ comparisons for the four tasks, respectively.
    $\dagger$ The screwing task was completed by Participant $1$ with the assistance of a carer, as it was a bimanual task.
    $\ddagger$ The cutting task is a bimanual task; Participant $1$ used their left residual limb (above elbow) to hold and manoeuvre the paper.
    }
    \vspace{-10pt}
    \label{fig:task_performance}
\end{figure*}

The flicking device was assessed with a devised disc flicking task, in which participants attempted to flick a disc at one of seven target discs, and hit rate was recorded as a performance metric; $35$ trials were performed in distance order---five attempts per target, from closest to farthest---after $15$ training trials. The flicking task was carried out on a flat whiteboard surface, using the 25 mm diameter by $8$~mm thick discs from the Jebsen-Taylor Hand Function Test. The target locations are arranged with a radial offset increasing by $100$~mm increments, and alternating $\pm 150$~mm axial offsets (see Fig.\ref{fig:task_performance} a).

The twisting device was assessed via a devised task in which participants screwed three screws into a test board --- a M4 panhead screw, a M3 Phillips screw, and a M2 Torx T6 screw, at depths of $8$~mm, $6$~mm, and $4$~mm, respectively. Completion time was recorded as the task performance metric.

The suction device was assessed with the timed stacking checkers task of the Jebsen-Taylor Hand Function Test~\cite{Jebsen1969AnFunction}, in which time to stack four checkers is recorded as the corresponding task performance metric.

The cutting device was evaluated via a cutting task, in which participants cut a template $10$~cm diameter circle from a sheet of paper, located approximately $25$~mm from the edge of the paper. Completion time and relative circularity were recorded as performance metrics. Circularity is defined as:
\begin{equation}
    \textit{circularity} = \frac{4\pi \times \textit{area}}{\textit{perimeter}^2},
\end{equation}
and has a maximum value of $1$. After task completion, circles were imaged and circularity calculated from extracted image contours. To account for image perspective, a template circle was simultaneously imaged and used as a reference circularity score. Relative circularity was computed as ratio of the circularities of the participant's cut circle to the template circle. Circularity was chosen over other metrics such as intersection-over-union because it is agnostic to the size of the cut shape; some participants chose to cut around the edge of the template, and some inside the template. It was reasoned that participants should not be penalised for this.

\subsection{Compensatory Motion}

Compensatory movement of the user was measured using four $6$~DOF Vive Trackers (HTC Ltd.) placed on the trunk, upper arm, and lower arm (see Fig. \ref{fig:evaluation} e), and a reference tracker placed on the desk at which the participant was sitting. Tracker pose information was streamed at $100$~Hz, with the $x$~axis of the trackers aligned with the axis of each body part that was tracked. Angular deviation of the three trackers from their initial orientation was computed as a proxy for compensatory movement. This was achieved by computing the relative rotation $R_{0\to t} = R_{t}R_{0}^{T}$ between the tracker's initial orientation $R_0$ and its orientation at time $t$, $R_t$.
The magnitude of this relative rotation, $\theta_t$, was computed:
\begin{equation}
    \theta_t = \bigg|\arccos\bigg(\frac{\Tr(R_{0\to t}) - 1}{2}\bigg)\bigg|.
\end{equation}
The mean absolute angle over the entire task was recorded. Computing the mean absolute angular deviation of each tracker is a less accurate, but simpler and more robust process than using motion capture and computing the joint angles of their arm and trunk, as in \cite{Vujaklija2023BiomechanicalTests}, and allowed us to recruit a relatively large number of participants to this study.

\subsection{User Perception}

To evaluate user perception we used the NASA Task Load Index (NASA-TLX) questionnaire~\cite{Hart1988DevelopmentResearch}, a validated method of subjectively evaluating mental workload~\cite{Rubio2004EvaluationMethods, Devos2020PsychometricAdults}, and here report the raw TLX scores of participants~\cite{Hart2006Nasa-TaskLater:} in order to obtain greater insight into task load influencing factors. In the NASA-TLX questionnaire, physical demand, mental demand, temporal demand, performance, effort, and frustration are self reported on a $20$-point scale (here displayed as a numerical value out of $100$ in increments of $5$).

\subsection{Humanoid Configuration}

{This subsection describes the control strategies used by participants} to complete each task with the OLYMPIC hand in its humanoid configuration {(see also Supplementary Videos S1-S4, which show examples of participants from both the humanoid and non-humanoid groups completing each task).}

\subsubsection{Flicking Task}

The passive extension springs of the index finger of the OLYMPIC hand were used to store and release flicking energy. Participants pushed the prosthesis against the flicking surface to flex the finger, then lifted the prosthesis off of the surface to release the finger.

\subsubsection{Screwing Task}

The screwdriver was held between the prosthesis' middle and ring fingers, and the position-controlled wrist pronation/supination DoF of the prosthesis was used to twist the screwdriver. The wrist position $q_{\text{wrist}}$ was controlled with a similar control scheme to the flicking device:
\begin{equation}
    q_{\text{wrist}}(t+\Delta t) = \begin{cases}
        q_{\text{wrist}}(t) + \bar{v}_{\text{screw}} \Delta t & s_1(t) - s_2(t) > \bar{s}_{\text{screw}} \\
        0 & s_2(t) - s_1(t) > \bar{s}_{\text{unscrew}} \\
        q_{\text{wrist}}(t) & \text{otherwise}
    \end{cases},
\end{equation}
with constant rotation rate of $\bar{v}_{\text{screw}}$, and screw and unscrew thresholds of $\bar{s}_{\text{screw}}$ and $\bar{s}_{\text{unscrew}}$, respectively.

\subsubsection{Stacking Task}

A binary control scheme was used to command the positions $q_{1:5}$ of all five fingers of the hand between their open $q_{\text{open}}$ and closed $q_{\text{close}}$ positions according to the differential signal from the two electrodes, with grasp and release thresholds of $\bar{s}_{\text{open}}$ and $\bar{s}_{\text{close}}$, respectively:
\begin{equation}
    q_{\text{1:5}} = \begin{cases}
        q_{\text{close}} & s_1(t) - s_2(t) > \bar{s}_{\text{close}} \\
        q_{\text{open}} & s_2(t) - s_1(t) > \bar{s}_{\text{open}} \\
    \end{cases}.
\end{equation}

\subsubsection{Cutting Task}

The prosthesis was not able to actuate the scissors, so the scissors were held in a neutral position and actuated by pressing the fingers of the hand against the desk.

\subsection{Statistical Analyses}

Mann-Whitney U tests were used to test for statistical significance between results from groups of participants, except for the binary hit or miss data from the flicking task, where paired proportion $z$-tests were used. Bonferroni corrections were used to account for multiple comparisons in all cases.

\section{Results and Discussion}\label{sec:results}

\subsection{Non-humanoid Terminal Devices Outperform Humanoid Terminal Devices}

The results of the four completed tasks can be seen in Fig.~\ref{fig:task_performance}, which highlight the effectiveness of the non-humanoid devices. Participants using the humanoid prosthesis were able to complete every task, demonstrating the versatility of humanoid prostheses. However, in all tasks, the humanoid group were dramatically outperformed by the non-humanoid group. It can be seen in Fig. \ref{fig:task_performance} \textbf{a} that the non-humanoid group had a hit rate $0.221$ higher than the humanoid group ($p<0.001$). On the screwing task (see Fig.~\ref{fig:task_performance} \textbf{b}), the non-humanoid group were significantly ($p<0.001$ for each trial) faster than the humanoid group, with average completion times of $278.6$ and $701.2$ seconds, respectively. The non-humanoid group completed the stacking checkers task significantly faster than the humanoid group ($p<0.01$ trials 1, 2, 4, 5, and $p<0.001$ trial 3) (see Fig.~\ref{fig:task_performance} \textbf{c}), with mean task completion times across all five trials of $30.3$ seconds and $77.1$ seconds, respectively. On the cutting task (see Fig.~\ref{fig:task_performance} \textbf{d}), the humanoid group significantly slower than the non-humanoid group ($p<0.001$) taking, on average, over four times as long to complete the task ($351.6$ seconds versus $83.2$ seconds). Furthermore, the circles cut by the non-humanoid group were significantly ($p<0.001$) more circular, with an average relative circularity of $0.91$ compared to $0.83$ for the humanoid group.

The task performance of the two participants with ULD generally aligned with those of the non-humanoid group. On the flicking task (Fig.~\ref{fig:task_performance}~\textbf{a}), Participant $1$ was somewhat better than the participant group, with an mean hit rate of $0.80$; possibly due to their considerable experience playing a similar flicking game prior to acquiring their limb difference. Participant $1$ is a bilateral amputee, with a transhumeral amputation on their left arm. This made the cutting task (Fig.~\ref{fig:task_performance}~\textbf{d}) difficult, as they used their left residuum to manoeuvre the paper during cutting (see Supplementary Video S4), and led to a longer completion time than the non-humanoid group, although still faster than the majority of the humanoid group. Similarly, the screwing task (Fig.~\ref{fig:task_performance} \textbf{b}) was impossible without a second hand, and was completed with carer assistance during screwdriver changes, resulting in a slightly faster mean completion time than the non-humanoid group of $237.6$ seconds. However, while these results are promising, due to the small sample size of participants with ULD, it cannot be guaranteed that they translate to the wider population of adults with ULD.

\begin{figure*}[!t]
    \centering
    \includegraphics[width=180mm]{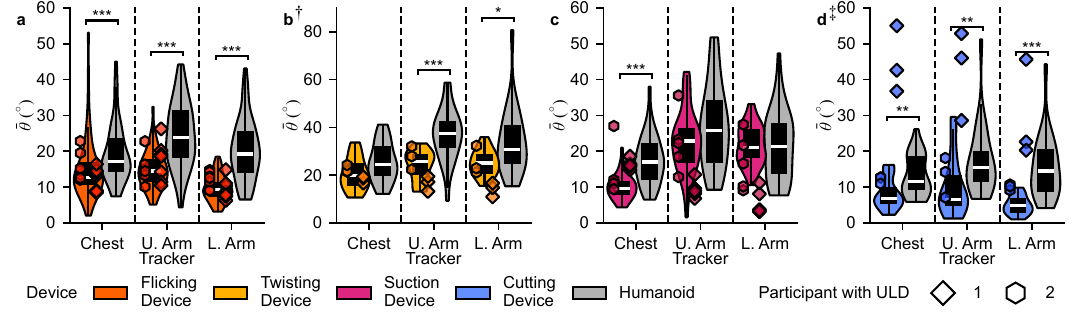}
    \vspace{-6pt}
    \caption{
    Mean absolute angular deviation $\bar{\theta}$ of trackers located on the chest, upper arm, and lower arm of groups of participants without ULD on the flicking task (\textbf{a}), screwing task (\textbf{b}), the stacking task (\textbf{c}), and the cutting task (\textbf{d}), with task performance of participants with ULD shown as individual points. Results of Mann-Whitney U tests: $^{*}p<0.05$, $^{**}p<0.01$, and $^{***}p<0.001$ after Bonferroni correction for 3 comparisons (\textbf{a-d}).
    $\dagger$ The screwing task was completed by Participant $1$ with the assistance of a carer, as it was a bimanual task.
    $\ddagger$ The cutting task is a bimanual task; Participant $1$ used their left residual limb (above elbow) to hold and manoeuvre the paper that was cut.
    }
    \vspace{-10pt}
    \label{fig:compensatory_motion}
\end{figure*}

\subsection{Non-humanoid Terminal Devices Reduce Compensatory Motion}

The absolute angular deviation of each body-mounted tracker for each task is shown in Fig.~\ref{fig:compensatory_motion} \textbf{a}-\textbf{d}. During the flicking task, because the humanoid group utilised the index finger extension springs to produce a flicking motion, a considerable amount of manoeuvring was required. This is visible in Fig.~\ref{fig:compensatory_motion} \textbf{a}, where the angular deviations of all trackers were significantly greater than the non-humanoid group ($p<0.001$ for all trackers). The non-humanoid group required very little arm motion to complete the screwing task. The humanoid group, due to an unavoidable radial offset of the screwdriver axis from the pronation/supination axis of the prosthesis, required considerably more compensatory motion to keep the tip of each screwdriver engaged with its respective screw (see Fig.~\ref{fig:compensatory_motion} \textbf{b}), and correspondingly, the angular deviation of the upper and lower arm trackers was significantly greater than for the non-humanoid group ($p<0.001$ and $p<0.05$, respectively). Shown in Fig. \ref{fig:compensatory_motion} \textbf{c}, the non-humanoid group utilized a significantly reduced amount of torso compensatory motion to complete the stacking task than the humanoid group ($p<0.001$), but exhibited similar levels of upper and lower arm angular deviation. This is because the humanoid group needed to drag the checkers away from the stacking platform so that they could be grasped, culminating in chest deviation while maintaining similar arm orientation. For the cutting task, the humanoid hand was not able to actuate the scissors --- the humanoid group manually pressed the scissors against the table to complete the task, resulting in a significant increase in angular deviation of all trackers ($p<0.01$ chest and upper arm, $p<0.001$ for lower arm), shown in Fig.~\ref{fig:compensatory_motion} \textbf{d}.

The results of the two participants with ULD generally align with the non-humanoid group. As mentioned previously, Participant $1$ used their left residuum to manoeuvre the paper during cutting and also used the table for further support. This resulted in a large increase in angular deviation of the chest, upper arm, and lower arm compared to the non-humanoid group (Fig.~\ref{fig:compensatory_motion} \textbf{d}). Participant $1$ completed the screwing task with carer assistance, meaning that there was a corresponding slight reduction in compensatory motion (Fig.~\ref{fig:compensatory_motion} \textbf{b}).

\subsection{Non-humanoid Prostheses Reduce Task Load}
\begin{figure*}[!t]
    \centering
    \includegraphics[width=180mm]{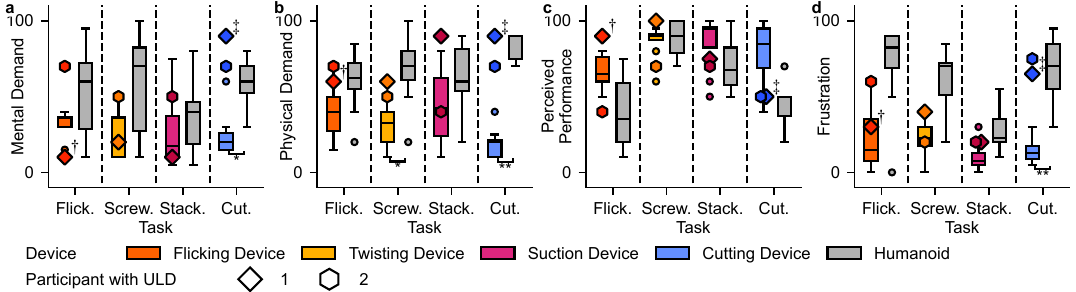}
    \vspace{-8pt}
    \caption{
    Raw TLX responses to \textbf{a} mental demand, \textbf{b} physical demand, \textbf{c} perceived performance, and \textbf{d} frustration for each device on their corresponding task. Flick. - flicking disc task, Screw. - screwing task, Stack. - stacking task, and Cut. - cutting task. All plots show results of Mann-Whitney U tests: $^{*}p<0.05$, $^{**}p<0.01$ after Bonferroni correction for 4 comparisons.
    $\dagger$ The screwing task was completed by Participant $1$ with the assistance of a carer, as it was a bimanual task.
    $\ddagger$ The cutting task is a bimanual task; Participant $1$ used their left residual limb (above elbow) to hold and manoeuvre the paper that was cut.
    }
    \vspace{-10pt}
    \label{fig:task_load}
\end{figure*}
Participant responses to the mental demand, physical demand, perceived performance, and frustration questions of the NASA-TLX questionnaire are shown in Fig.~\ref{fig:task_load}. Mental demand (Fig.~\ref{fig:task_load} \textbf{a}) was significant lower for the non-humanoid group on the cutting task ($p<0.05$). Physical demand (see Fig.~\ref{fig:task_load} \textbf{b}) was significantly lower for the non-humanoid groups on the screwing and cutting tasks ($p<0.05$ and $p<0.01$, respectively). This corresponds to the tasks with the largest differences in completion times between groups. In terms of perceived performance (see Fig.~\ref{fig:task_load} \textbf{c}), there were no significant differences between the humanoid and non-humanoid groups for any of the tasks. Interestingly, for both groups, the flicking and cutting tasks had the lowest perceived performance, possibly due to the visible performance cues present in these tasks: hit success and circle roundness, respectively. Finally, frustration was significantly lower ($p<0.01$) for the non-humanoid group on the cutting task (see Fig.~\ref{fig:task_load} \textbf{d}). Generally, the non-humanoid groups had lower median mental demand, physical demand, and frustration, and higher median perceived performance, but further testing with a larger sample size would be needed to verify that these trends are significant.

For participants with ULD, physical demand was generally higher than those without ULD in the non-humanoid group. This may be due to both atrophied muscles used in myoelectric control and the concentrated prosthesis weight on the participant's residual limb. As expected, the cutting task elicited greater mental demand and frustration, and reduced perceived performance for Participant $1$, due to the complexities of completing the task as a bilateral amputee. For Participant $2$, frustration was strongly related to perceived performance and mental demand, evidenced by high frustration for the flicking and cutting tasks. It should be noted that participants with ULD completed the tasks in fixed order, and as such ordering effects may be present, particularly in task load results. A larger sample size, and counterbalanced or randomised task order would be necessary to counteract this.

\section{Conclusions \& Future Work}\label{sec:conclusions}

Four terminal devices were developed as illustrative examples of task-specific electromechanical prostheses, and, as expected for terminal devices specifically designed for each evaluation task, in all cases outperformed the humanoid control prosthesis. The terminal devices allowed participants to complete timed tasks---the screwing, stacking checkers, and cutting tasks---in a fraction of the time of the humanoid group, and precision tasks --- the flicking and cutting tasks -- with a higher accuracy than the humanoid group, contributing to significant reductions in mental demand, physical demand, and frustration. These benefits translated from participants without to participants with ULD, but special attention must be paid to task load to foster a sense of achievement and minimise frustration, and overexertion should be avoided wherever possible. Practically, tasks such as cutting and screwing go from being impractical with a conventional prosthesis to within a user's capabilities, offering hope that functional non-humanoid prostheses could aid users in returning to work, particularly in physically-skilled jobs. Overall, these results are encouraging, given the pressing need for improved prosthesis functionality \cite{Biddiss2009ConsumerProsthetics, Espinosa2019UnderstandingAbandonment} and the long-term health complications that arise from reliance on compensatory motion to operate a prosthesis~\cite{Ephraim2005PhantomSurvey, Hanley2009ChronicLoss}. Furthermore, the flicking device shines a light on an overlooked area of prosthesis design: social tasks. With lower reported life satisfaction \cite{stlie2011MentalGroup} and social functioning \cite{Johansen2016Health-relatedStudy} in adults with ULD, prostheses designed for social tasks could play a large role in improving the lives of this population.

We anticipate that this study could lead to a change in prosthesis design, leading to functional prostheses that empower users to perform physically and socially meaningful actions. Due to limitations in existing prostheses, simpler devices that provide such a substantial boost in functionality, usability, and long-term comfort are not only a viable option; they are a favourable one. This work opens the door to a host of exciting and impactful future research, from functional prosthesis design to investigation into haptic feedback from and the psychological embodiment of non-humanoid prostheses. An open question remains regarding the most effective mechanical implementation of such prostheses. In this work, the OLYMPIC hand's modular finger design~\cite{Liow2020OLYMPIC:Mechanisms} was leveraged, which would allow users to mount a versatile set of chosen terminal devices, minimising the need to repeatedly interchange end effectors and providing users with the freedom of choice in creating their own personalised prosthesis. However, we acknowledge that maintaining a stock of attachable terminal devices may be inconvenient; it may be the case that terminal devices could be grouped to perform subsets of commonly associated tasks. Finally, the devices developed in this study are all single DOF, and higher DOF devices could be developed to perform more complex actions, such as planar writing motion.


\section*{Acknowledgements}
We would like to thank all participants for taking part in this study, the Alex Lewis Trust for their support, and C. Traweek, H. Young, K. Li, M. Boguslavskiy, W. Chen, X. Wang, P. Slade, and T. Nanayakkara for their feedback on this work, and Y. Ge and D. Lee for assistance in facilitating the study.

\bibliographystyle{ieeetr}
\bibliography{references.bib}

\begin{thebibliography}{10}

\bibitem{McDonald2021GlobalAmputation}
C.~L. McDonald, S.~Westcott-Mccoy, M.~R. Weaver, J.~Haagsma, and D.~Kartin, ``{Global prevalence of traumatic non-fatal limb amputation},'' {\em Prosthetics and Orthotics International}, vol.~45, pp.~105--114, 4 2021.

\bibitem{Battye1955TheProstheses.}
C.~K. Battye, A.~Nightingale, and J.~Whillis, ``{The use of myo-electric currents in the operation of prostheses.},'' {\em The Journal of bone and joint surgery. British volume}, vol.~37 B, pp.~506--510, 8 1955.

\bibitem{Kyberd1994TheProsthesis.}
P.~J. Kyberd and P.~H. Chappell, ``{The Southampton Hand: an intelligent myoelectric prosthesis.},'' {\em Journal of Rehabilitation Research and Development}, vol.~31, pp.~326--334, 11 1994.

\bibitem{Weiner2018TheControl}
P.~Weiner, J.~Starke, F.~Hundhausen, J.~Beil, and T.~Asfour, ``{The KIT Prosthetic Hand: Design and Control},'' {\em IEEE International Conference on Intelligent Robots and Systems}, pp.~3328--3334, 12 2018.

\bibitem{Kerver2023ThePromise}
N.~Kerver, V.~Schuurmans, C.~K. van~der Sluis, and R.~M. Bongers, ``{The multi-grip and standard myoelectric hand prosthesis compared: does the multi-grip hand live up to its promise?},'' {\em Journal of NeuroEngineering and Rehabilitation 2023 20:1}, vol.~20, pp.~1--18, 2 2023.

\bibitem{Kyberd2007SurveyKingdom}
P.~J. Kyberd, C.~Wartenberg, L.~Sandsj{\"{o}}, S.~J{\"{o}}nsson, D.~Gow, J.~Frid, C.~Almstr{\"{o}}m, and L.~Sperling, ``{Survey of upper-extremity prosthesis users in Sweden and the United Kingdom},'' {\em Journal of Prosthetics and Orthotics}, vol.~19, pp.~55--62, 4 2007.

\bibitem{Pylatiuk2007ResultsUsers}
C.~Pylatiuk, S.~Schulz, and L.~D{\"{o}}derlein, ``{Results of an internet survey of myoelectric prosthetic hand users},'' {\em Prosthetics and Orthotics International}, vol.~31, pp.~362--370, 12 2007.

\bibitem{stlie2012ProsthesisSurvey}
K.~{\O}stlie, I.~M. Lesj{\o}, R.~J. Franklin, B.~Garfelt, O.~H. Skjeldal, and P.~Magnus, ``{Prosthesis rejection in acquired major upper-limb amputees: a population-based survey},'' {\em Disability and Rehabilitation: Assistive Technology}, vol.~7, pp.~294--303, 7 2012.

\bibitem{Salminger2020CurrentAcceptance}
S.~Salminger, H.~Stino, L.~H. Pichler, C.~Gstoettner, A.~Sturma, J.~A. Mayer, M.~Szivak, and O.~C. Aszmann, ``{Current rates of prosthetic usage in upper-limb amputees – have innovations had an impact on device acceptance?},'' {\em Disability and Rehabilitation}, vol.~44, no.~14, pp.~3708--3713, 2020.

\bibitem{Biddiss2009ConsumerProsthetics}
E.~Biddiss, D.~Beaton, and T.~Chau, ``{Consumer design priorities for upper limb prosthetics},'' {\em Disability and Rehabilitation: Assistive Technology}, vol.~2, no.~6, pp.~346--357, 2009.

\bibitem{Espinosa2019UnderstandingAbandonment}
M.~Espinosa and D.~Nathan-Roberts, ``{Understanding Prosthetic Abandonment},'' {\em Proceedings of the Human Factors and Ergonomics Society Annual Meeting}, vol.~63, pp.~1644--1648, 11 2019.

\bibitem{Stephens-Fripp2019AHands}
B.~Stephens-Fripp, M.~Jean~Walker, E.~Goddard, and G.~Alici, ``{A survey on what Australians with upper limb difference want in a prosthesis: justification for using soft robotics and additive manufacturing for customized prosthetic hands},'' {\em Disability and Rehabilitation: Assistive Technology}, vol.~15, pp.~342--349, 4 2019.

\bibitem{Yamamoto2019Cross-sectionalProstheses}
M.~Yamamoto, K.~C. Chung, J.~Sterbenz, M.~J. Shauver, H.~Tanaka, T.~Nakamura, J.~Oba, T.~Chin, and H.~Hirata, ``{Cross-sectional International Multicenter Study on Quality of Life and Reasons for Abandonment of Upper Limb Prostheses},'' {\em Plastic and Reconstructive Surgery - Global Open}, vol.~7, p.~E2205, 5 2019.

\bibitem{Datta2004FunctionalCongenital}
D.~Datta, K.~Selvarajah, and N.~Davey, ``{Functional outcome of patients with proximal upper limb deficiency–acquired and congenital},'' {\em Clinical Rehabilitation}, vol.~18, pp.~172--177, 3 2004.

\bibitem{vanderSluis2009JobAmputees}
C.~K. van~der Sluis, P.~P. Hartman, T.~Schoppen, and P.~U. Dijkstra, ``{Job Adjustments, Job Satisfaction and Health Experience in Upper and Lower Limb Amputees},'' {\em Prosthetics and Orthotics International}, vol.~33, pp.~41--51, 1 2009.

\bibitem{Postema2016UpperProductivity}
S.~G. Postema, R.~M. Bongers, M.~A. Brouwers, H.~Burger, L.~M. Norling-Hermansson, M.~F. Reneman, P.~U. Dijkstra, and C.~K. van~der Sluis, ``{Upper Limb Absence: Predictors of Work Participation and Work Productivity},'' {\em Archives of Physical Medicine and Rehabilitation}, vol.~97, pp.~892--899, 6 2016.

\bibitem{Liow2020OLYMPIC:Mechanisms}
L.~Liow, A.~B. Clark, and N.~Rojas, ``{OLYMPIC: A modular, tendon-driven prosthetic hand with novel finger and wrist coupling mechanisms},'' {\em IEEE Robotics and Automation Letters}, vol.~5, pp.~299--306, 4 2020.

\bibitem{Johannes2020TheLimb}
M.~S. Johannes, E.~L. Faulring, K.~D. Katyal, M.~P. Para, J.~B. Helder, A.~Makhlin, T.~Moyer, D.~Wahl, J.~Solberg, S.~Clark, R.~S. Armiger, T.~Lontz, K.~Geberth, C.~W. Moran, B.~A. Wester, T.~Van~Doren, and J.~J. Santos-Munne, ``{The Modular Prosthetic Limb},'' {\em Wearable Robotics: Systems and Applications}, pp.~393--444, 1 2020.

\bibitem{Hahne2018_LinearRegression2DOF}
J.~M. Hahne, M.~A. Schweisfurth, M.~Koppe, and D.~Farina, ``{Simultaneous control of multiple functions of bionic hand prostheses: Performance and robustness in end users},'' {\em Science Robotics}, vol.~3, 6 2018.

\bibitem{Dyson2018MyoelectricDecoders}
M.~Dyson, J.~Barnes, and K.~Nazarpour, ``{Myoelectric control with abstract decoders},'' {\em Journal of Neural Engineering}, vol.~15, p.~056003, 7 2018.

\bibitem{Dyson2020LearningControl}
M.~Dyson, S.~Dupan, H.~Jones, and K.~Nazarpour, ``{Learning, Generalization, and Scalability of Abstract Myoelectric Control},'' {\em IEEE Transactions on Neural Systems and Rehabilitation Engineering}, vol.~28, pp.~1539--1547, 7 2020.

\bibitem{Dollar2007TheStudy}
A.~M. Dollar and R.~D. Howe, ``{The SDM hand as a prosthetic terminal device: A feasibility study},'' {\em 2007 IEEE 10th International Conference on Rehabilitation Robotics, ICORR'07}, pp.~978--983, 2007.

\bibitem{Cheng2016ProstheticRobotics}
N.~Cheng, J.~Amend, T.~Farrell, D.~Latour, C.~Martinez, J.~Johansson, A.~McNicoll, M.~Wartenberg, S.~Naseef, W.~Hanson, and W.~Culley, ``{Prosthetic Jamming Terminal Device: A Case Study of Untethered Soft Robotics},'' {\em Soft Robotics}, vol.~3, pp.~205--212, 12 2016.

\bibitem{Yoshikawa2023Finch:Bulge}
M.~Yoshikawa, K.~Ogawa, S.~Yamanaka, and N.~Kawashima, ``{Finch: Prosthetic Arm With Three Opposing Fingers Controlled by a Muscle Bulge},'' {\em IEEE Transactions on Neural Systems and Rehabilitation Engineering}, vol.~31, pp.~377--386, 2023.

\bibitem{Hong2023Angle-programmedUltraprecision}
Y.~Hong, Y.~Zhao, J.~Berman, Y.~Chi, Y.~Li, H.~H. Huang, and J.~Yin, ``{Angle-programmed tendril-like trajectories enable a multifunctional gripper with ultradelicacy, ultrastrength, and ultraprecision},'' {\em Nature Communications 2023 14:1}, vol.~14, pp.~1--10, 8 2023.

\bibitem{Kyberd2021MakingArms}
P.~J. Kyberd, ``{Making Hands: A History of Prosthetic Arms},'' {\em Making Hands: A History of Prosthetic Arms}, pp.~1--364, 1 2021.

\bibitem{Highsmith2007KinematicAmputation}
M.~J. Highsmith, S.~L. Carey, K.~W. Koelsch, C.~P. Lusk, and M.~E. Maitland, ``{Kinematic evaluation of terminal devices for kayaking with upper extremity amputation},'' {\em Journal of Prosthetics and Orthotics}, vol.~19, pp.~84--90, 7 2007.

\bibitem{Hammond2012TowardsHands}
F.~L. Hammond, J.~Weisz, A.~A. De~La Llera~Kurth, P.~K. Allen, and R.~D. Howe, ``{Towards a design optimization method for reducing the mechanical complexity of underactuated robotic hands},'' {\em IEEE International Conference on Robotics and Automation}, pp.~2843--2850, 2012.

\bibitem{Maat2018PassiveReview}
B.~Maat, G.~Smit, D.~Plettenburg, and P.~Breedveld, ``{Passive prosthetic hands and tools: A literature review},'' {\em Prosthetics and Orthotics International}, vol.~42, pp.~66--74, 2 2018.

\bibitem{Smit2010EfficiencyProstheses}
G.~Smit and D.~H. Plettenburg, ``{Efficiency of voluntary closing hand and hook prostheses},'' {\em Prosthetics and Orthotics International}, vol.~34, pp.~411--427, 12 2010.

\bibitem{Smit2012EfficiencyDevelopment}
G.~Smit, R.~M. Bongers, C.~K. Van~der Sluis, and D.~H. Plettenburg, ``{Efficiency of voluntary opening hand and hook prosthetic devices: 24 years of development?},'' {\em Journal of rehabilitation research and development}, vol.~49, pp.~523--534, 6 2012.

\bibitem{Bragaru2012SportLiterature}
M.~Bragaru, R.~Dekker, and J.~H. Geertzen, ``{Sport prostheses and prosthetic adaptations for the upper and lower limb amputees: an overview of peer reviewed literature},'' {\em Prosthetics and Orthotics International}, vol.~36, pp.~290--296, 8 2012.

\bibitem{Walker2020TowardsDifference}
M.~J. Walker, E.~Goddard, B.~Stephens-Fripp, and G.~Alici, ``{Towards Including End-Users in the Design of Prosthetic Hands: Ethical Analysis of a Survey of Australians with Upper-Limb Difference},'' {\em Science and Engineering Ethics}, vol.~26, pp.~981--1007, 4 2020.

\bibitem{Arabian2016GlobalDevices}
A.~Arabian, D.~Varotsis, C.~McDonnell, and E.~Meeks, ``{Global social acceptance of prosthetic devices},'' {\em GHTC 2016 - IEEE Global Humanitarian Technology Conference: Technology for the Benefit of Humanity, Conference Proceedings}, pp.~563--568, 2016.

\bibitem{Jiang2014AHand}
L.~Jiang, B.~Zeng, S.~Fan, K.~Sun, T.~Zhang, and H.~Liu, ``{A modular multisensory prosthetic hand},'' {\em IEEE International Conference on Information and Automation}, pp.~648--653, 10 2014.

\bibitem{Johansson1987SignalsGrip}
R.~S. Johansson and G.~Westling, ``{Signals in tactile afferents from the fingers eliciting adaptive motor responses during precision grip},'' {\em Experimental Brain Research}, vol.~66, pp.~141--154, 3 1987.

\bibitem{Chappell2022TowardsDistance}
D.~Chappell, Z.~Yang, H.~W. Son, F.~Bello, P.~Kormushev, and N.~Rojas, ``{Towards Instant Calibration in Myoelectric Prosthetic Hands: A Highly Data-Efficient Controller Based on the Wasserstein Distance},'' in {\em IEEE International Conference on Rehabilitation Robotics (ICORR)}, (Rotterdam), pp.~1--6, IEEE, 7 2022.

\bibitem{Resnik2018HowProstheses}
L.~J. Resnik, M.~L. Borgia, F.~Acluche, J.~M. Cancio, G.~Latlief, and N.~Sasson, ``{How do the outcomes of the DEKA Arm compare to conventional prostheses?},'' {\em PLOS ONE}, vol.~13, p.~e0191326, 1 2018.

\bibitem{Mohammadi2022PreliminaryProsthesis}
A.~Mohammadi, J.~Lavranos, Y.~Tan, P.~Choong, and D.~Oetomo, ``{Preliminary Clinical Evaluation of the X-Limb Hand: A 3D Printed Soft Robotic Hand Prosthesis},'' {\em Biosystems and Biorobotics}, vol.~28, pp.~869--873, 2022.

\bibitem{Jebsen1969AnFunction}
R.~H. Jebsen, N.~Taylor, R.~B. Trieschmann, M.~J. Trotter, and L.~A. Howard, ``{An objective and standardized test of hand function},'' {\em Arch. Phys. Med. Rehabil.}, vol.~50, no.~6, pp.~311--319, 1969.

\bibitem{Trombly1983OccupationalDysfunction}
C.~A. Trombly and A.~D. Scott, {\em {Occupational Therapy for Physical Dysfunction}}.
\newblock Baltimore: Williams and Wilkins, 2~ed., 1983.

\bibitem{Light2002SHAP}
C.~M. Light, P.~H. Chappell, and P.~J. Kyberd, ``{Establishing a standardized clinical assessment tool of pathologic and prosthetic hand function: Normative data, reliability, and validity},'' {\em Archives of Physical Medicine and Rehabilitation}, vol.~83, pp.~776--783, 6 2002.

\bibitem{Tiffin1948TheValidity.}
J.~Tiffin and E.~J. Asher, ``{The Purdue Pegboard: norms and studies of reliability and validity.},'' {\em Journal of applied psychology}, vol.~32, no.~3, p.~234, 1948.

\bibitem{Chappell2022VirtualInteraction}
D.~Chappell, H.~W. Son, A.~B. Clark, Z.~Yang, F.~Bello, P.~Kormushev, and N.~Rojas, ``{Virtual Reality Pre-Prosthetic Hand Training with Physics Simulation and Robotic Force Interaction},'' {\em IEEE Robotics and Automation Letters}, vol.~7, pp.~4550--4557, 4 2022.

\bibitem{Parr2023APROS-TLX}
J.~V. Parr, A.~Galpin, L.~Uiga, B.~Marshall, D.~J. Wright, Z.~C. Franklin, and G.~Wood, ``{A tool for measuring mental workload during prosthesis use: The Prosthesis Task Load Index (PROS-TLX)},'' {\em PLOS ONE}, vol.~18, p.~e0285382, 5 2023.

\bibitem{Hunt2017Pham:Measure}
C.~Hunt, R.~Yerrabelli, C.~Clancy, L.~Osborn, R.~Kaliki, and N.~Thakor, ``{Pham: prosthetic hand assessment measure},'' in {\em Proceedings of the Myoelectric Control Symposium (MEC17)}, pp.~221--224, 2017.

\bibitem{Vujaklija2023BiomechanicalTests}
I.~Vujaklija, M.~K. Jung, T.~Hasenoehrl, A.~D. Roche, A.~Sturma, S.~Muceli, R.~Crevenna, O.~C. Aszmann, and D.~Farina, ``{Biomechanical Analysis of Body Movements of Myoelectric Prosthesis Users During Standardized Clinical Tests},'' {\em IEEE Transactions on Biomedical Engineering}, vol.~70, pp.~789--799, 3 2023.

\bibitem{Hart1988DevelopmentResearch}
S.~G. Hart and L.~E. Staveland, ``{Development of NASA-TLX (Task Load Index): Results of Empirical and Theoretical Research},'' {\em Advances in Psychology}, vol.~52, pp.~139--183, 1 1988.

\bibitem{Rubio2004EvaluationMethods}
S.~Rubio, E.~D{\'{i}}az, J.~Mart{\'{i}}n, and J.~M. Puente, ``{Evaluation of Subjective Mental Workload: A Comparison of SWAT, NASA-TLX, and Workload Profile Methods},'' {\em Applied Psychology}, vol.~53, pp.~61--86, 1 2004.

\bibitem{Devos2020PsychometricAdults}
H.~Devos, K.~Gustafson, P.~Ahmadnezhad, K.~Liao, J.~D. Mahnken, W.~M. Brooks, and J.~M. Burns, ``{Psychometric Properties of NASA-TLX and Index of Cognitive Activity as Measures of Cognitive Workload in Older Adults},'' {\em Brain Sciences 2020, Vol. 10, Page 994}, vol.~10, p.~994, 12 2020.

\bibitem{Hart2006Nasa-TaskLater:}
S.~G. Hart, ``{Nasa-Task Load Index (NASA-TLX); 20 Years Later:},'' {\em Proceedings of the Human Factors and Ergonomics Society Annual Meeting}, vol.~50, pp.~904--908, 11 2006.

\bibitem{Ephraim2005PhantomSurvey}
P.~L. Ephraim, S.~T. Wegener, E.~J. MacKenzie, T.~R. Dillingham, and L.~E. Pezzin, ``{Phantom Pain, Residual Limb Pain, and Back Pain in Amputees: Results of a National Survey},'' {\em Archives of Physical Medicine and Rehabilitation}, vol.~86, pp.~1910--1919, 10 2005.

\bibitem{Hanley2009ChronicLoss}
M.~A. Hanley, D.~M. Ehde, M.~Jensen, J.~Czerniecki, D.~G. Smith, and L.~R. Robinson, ``{Chronic Pain Associated with Upper-Limb Loss},'' {\em American Journal of Physical Medicine {\&} Rehabilitation}, vol.~88, p.~742, 8 2009.

\bibitem{stlie2011MentalGroup}
K.~{\O}stlie, P.~Magnus, O.~H. Skjeldal, B.~Garfelt, and K.~Tambs, ``{Mental health and satisfaction with life among upper limb amputees: a Norwegian population-based survey comparing adult acquired major upper limb amputees with a control group},'' {\em Disability and Rehabilitation}, vol.~33, no.~17-18, pp.~1594--1607, 2011.

\bibitem{Johansen2016Health-relatedStudy}
H.~Johansen, K.~{\O}stlie, L.~{\O}. Andersen, and S.~Rand-Hendriksen, ``{Health-related quality of life in adults with congenital unilateral upper limb deficiency in Norway. A cross-sectional study},'' {\em Disability and Rehabilitation}, vol.~38, pp.~2305--2314, 11 2016.

\end{thebibliography}

\end{document}